\documentclass[conference]{IEEEtran}
\IEEEoverridecommandlockouts
\usepackage{cite}
\usepackage{amsmath,amssymb,amsfonts}
\usepackage{algorithmic}
\usepackage{graphicx}
\usepackage{textcomp}
\usepackage{multirow}
\usepackage{xcolor}
\usepackage{amsthm}
\usepackage{adjustbox}
\usepackage[english]{babel}
\usepackage[T1]{fontenc}
\usepackage[utf8]{inputenc}
\usepackage{multirow,siunitx}
\usepackage{arydshln}

\def\BibTeX{{\rm B\kern-.05em{\sc i\kern-.025em b}\kern-.08em
    T\kern-.1667em\lower.7ex\hbox{E}\kern-.125emX}}
\begin{document}

\author{\IEEEauthorblockN{Amir Jalilifard\IEEEauthorrefmark{1},
Dehua Chen\IEEEauthorrefmark{1},
Lucas Pereira Lopes\IEEEauthorrefmark{1},
Isaac Ben-Akiva\IEEEauthorrefmark{1},
Pedro Henrique Gonçalves Inazawa\IEEEauthorrefmark{1}
}
\IEEEauthorblockA{\IEEEauthorrefmark{1}PicPay Behavioral Science Team\\}
}

\title{Friendship is All we Need: A Multi-graph Embedding Approach for Modeling Customer Behavior}

\maketitle

\begin{abstract}
Understanding customer behavior is fundamental for many use-cases in industry, especially in accelerated growth areas such as fin-tech and e-commerce. Structured data are often expensive, time-consuming and inadequate to analyze and study complex customer behaviors. In this paper, we propose a multi-graph embedding approach for creating a non-linear representation of customers in order to have a better knowledge of their characteristics without having any prior information about their financial status or their interests. By applying the current method we are able to predict users' future behavior with a reasonably high accuracy only by having the information of their friendship network. Potential applications include recommendation systems and credit risk forecasting.
\end{abstract}


\section{Introduction} \label{sec:int}
Better knowledge of user's financial affordability and his general characteristics, and predicting the customer's future behavior based on these information makes it possible to efficiently direct resources at sales and marketing departments. This also helps offer the best services and products that fit user's interests and provide more detailed information about creditworthiness of customers. Although structured information such as those from credit reporting agencies/bureaus and transaction/consumption historical activities are helpful in understanding user's behavior pattern and his creditworthiness, these type of information is expensive, time-consuming to collect and can contain some issues \cite{avery2004credit}.
Moreover, structured data like credit bureaus and transaction information, if exist at all, do not always provide an accurate insight and correct information regarding customer's characteristics. On the other hand, since people's behavior are mainly influenced by several complex mental and social processes, traditional data sources are not able to reflect complex decision making process made by customer \cite{avery2004credit}, \cite{scurlock2007maxed}.

Inspired by social support theory, social media concept and relationship quality on social networks \cite{vaux1988social}, \cite{hajli2014role},\cite{amatulli2014does},\cite{harris2011engaging}, this study aims at investigating the possibility of learning users' behavior by understanding their nearby friends behaviours in embedding space. Therefore, we propose using friendship networks in order to infer user's financial credibility and understanding his buying behavior without having any direct knowledge about his financial or personal records. 

We briefly introduce our approach here. Firstly, we create a vector representation of buying habits and creditworthiness of a limited group of users. Then we infer the financial affordability and buying habits of those who are somehow in their friendship network. We continually use the latest updated group of nodes to infer the new status of their connected components in a graph.

The main contributions of our works are stated as follows:
\begin{itemize}
    \item We showed that it is possible to achieve a reliable prediction results without directly using the information of users by knowing their friends.
    \item Our approach is effective in several different applications, such as predicting user consuming activities and evaluating credit worthiness.
    \item We presented a method that is able to generate a representation which is close to real representation of unseen users.
\end{itemize}

This paper is organized as follows: in section \ref{sec:relwork}, we review the related works. Section~\ref{sec:data} provides a detailed description of the data utilized in the current study. In section ~\ref{sec:learning} a background information about graph representation learning is presented and the proposed multi-graph architecture, designed to generate user embeddings, is explained. In section \ref{sec:res} we present the results which is then followed by conclusion in Section~\ref{sec:con}.

\section{Related Works} \label{sec:relwork}

While customer analysis has been an interesting area for a long time in behavior research \cite{herniter1971probablistic}, recently using big data and predictive models for modeling customer behaviors and credit risk has received a lot of attention~\cite{kruppa2013consumer, martinez2020machine, alvino2018towards}. Kachamas et al. developed a social media analysis tool which is able analyze customer opinion about products by utilizing data collected from Facebook comments \cite{kachamas2019application}. This method measures sentiment toward products or services in form of positive, negative and neutral opinions. Afterward, it analyzes messages posted by brands communicating to customer in order to predict behavior and to determine whether it follows online consumer behavior based on Dentsu's AISAS model \cite{sugiyama2011dentsu}. Badea used a group of features such as credit card information, customer region and employment data, fed them to a Neural Networks and showed that it has a superior performance in comparison with conventional methods like discriminant analysis \cite{badea2014predicting}. Lang et al. studied the possibility of applying Recurrent Neural Networks on a sequence of user's online activities (i.e. ad-click, product view, card addition, etc) to model customer behavior.

Credit scoring is often considered a binary classification problem where the creditworthiness of a customer is either good or bad, although this worthiness can also be demonstrated with probabilities. Analysing credit risk using machine learning and based on payment history and credit bureaus have been considered in many studies \cite{kruppa2013consumer}, \cite{khandani2010consumer}, \cite{kalayci2018credit}. Independent of algorithm applied to predict users' credit risk, all approaches tend to use all or some of the following data: a) capacity which is the ability of the borrower to pay the loan; b) capital structure which reflects the companies liability; c) coverage which is defined as a mean of protection in case of loan loss if default happens; d) character that is a history of borrower's payments, committed frauds and legal expenses; and e) conditions which are the circumstances that are out of loaner's control (i.e. inflation rate, unemployment, etc). One shortcoming of these types of data is that they are hard to obtain, time consuming to be updated and expensive to buy due to their diverse nature. Furthermore, these data are only available for users with several years of payment history which makes it useless for customers without or with insufficient financial records. 

Graphs are important structures that are used to model a set of objects as a source of knowledge that in some sense are related. Although vastly employed in a variety of applications such as molecular structure, digital maps and time-series data, recently, graphs have been considered as an interesting source of pattern recognition by researchers. One may be interested in learning the structure of friendships in social networks and suggesting new friends for users, or modeling and predicting drug side-effects by learning drug-protein relation in a graph. No matter which application of graph learning we are interested in, the major challenge is to find a way to incorporate the complex structure of graphs (nodes and edges) into the existing machine learning algorithms. In order to address this challenge several models where proposed aiming at learning the structure and creating a lower dimensional representation of graphs (called embedding) using machine learning algorithms \cite{angles2008survey}, \cite{belkin2002laplacian}, \cite{grover2016node2vec} and \cite{kipf2016semi}.

Learning user behavior in large networks and predicting his possible interests is one of the applications of graph embedding \cite{goyal2018graph}. Yuana and colleagues utilized heterogeneous information, including user attributes, user
network structure, user network connection, and user behavior labels in order to learn user semantic properties through embeddings and to predict user behaviors. Recently, some researchers considered creating global user representation based on integration of multiple information from different sources (sites, social networks, etc) \cite{buccafurri2016model}, \cite{shi2015semantic}. Zhang and Yu proposed an unsupervised method to discover the potential links between users and between locations for multiple source \cite{shi2015semantic}. Yuan et al. considered using multiple separate information from multiple social networks to create multiple heterogeneous graphs and to learn the node representations separately \cite{yuan2019user}. subsequently, they apply a weighted average to form the final representation of user-user, user-item and item-item nodes and then used these embeddings to recommend friends and items. Although this study concentrates on creating global embeddings for user behavior by utilizing multiple sources and through complex weighting process, these global representations have shown poor performance in recommending relevant users and items (see the results in \cite{yuan2019user}).

In the next section the data structure and methods of the current study are demonstrated. We first go through the details of the data structure and then describe the learning strategy.

\section{Data} \label{sec:data}
We used three sources of data: a) friendship relations among more than two million users; b) more than four millions user transactions registered in our database; and 3) data of more than one hundred thousand sellers and their attributes such as their names, market segment and location features. Afterwards, we built a graph for each source of data, resulting in two heterogeneous graphs, user transactions (user-seller) and seller attributes (seller-attribute), and one homogeneous graphs, friendship (user-user). All of these graphs usually exhibit a sparse structure. We utilize the structural roles of the friendship, buying habits and commercial sellers' attributes in order to learn three types of representations for user-user, user-seller and seller-attributes interactions. In order to generate more informative attributes for seller's features, keywords from sellers' names and their occupation descriptions, considering the frequency of words, were extracted. 

We apply SkipGram~\cite{grover2016node2vec} to learn representations of these graphs in an unsupervised manner, but instead of performing expensive random walk-like context, we employ a restricted number of permutations over the immediate neighborhood of a node as context to generate its representation \cite{pimentel2018fast}. Further, we unify multi-graph relations in order to create a novel concept defined as buying habits (and possible future buying habits) of friends and those who have a high probability of making friendship with user in the future.

\section{Learning Customer Representations} \label{sec:learning}
The context of a word is usually approximated by the words surrounding it. In graphs, a node's context is an even more complex concept. DeepWalk \cite{perozzi2014deepwalk} and Node2Vec use random walks of size \textit{l} to simulate the context with window size \textit{l} in SkipGram. In contrast, the context of a node is based on the neighborhoods of nodes, defined as the direct connection between nodes. Consequently, nodes' representations will be mainly defined by their neighborhoods, and those nodes with similar neighborhoods will be associated with similar representations. This results in embeddings focused mainly on the second-order proximity. Neighborhood Based Node Embeddings (NBNE) separates a nodes' neighborhood in small chunks and then maximizes the log likelihood of predicting a node given another in the same chunk \cite{pimentel2018fast}.

NBNE makes clusters of nodes based on their neighborhoods. There are two main challenges in forming groups from neighborhoods, as follows:

\begin{itemize}
\item Groups based in neighbors are complex because nodes have different degrees.
\item There is no explicit way to choose the order in which they would appear in a group, unlike text, where the order is clear.
\end{itemize}

To deal with these challenges, NBNE uses random permutations to form small chunks with only $K$ neighbors. This number of permutations $n$ controls the trade-off between training time and increasing the training dataset. Selecting a higher value for $n$ results in a more uniform distribution on possible neighborhood groups, but it also increases training time.

\subsection{Learning Representations} \label{subsec:learning_representations}

NBNE creates a set of groups $S$, where each member of $S$ is a node from the chuck. It then trains the vector representations of nodes by maximizing the log likelihood to predict a node given another node in a chunk and given a set of representations $r$, making each node in a chunk predict the rest of the nodes in the same chunk. The log likelihood maximized by NBNE is given by:
\begin{equation} \label{sg_entropy}
  \max\limits_{r} \quad \frac{1}{|S|}  \sum_{s \in S} \left( \log \left( p\left( s | r \right) \right)\right)
\end{equation}
\noindent where $p\left( s | r \right)$ is the probability of each group, which is given by:
\begin{equation} \label{eq:sentence_prob}
  \log \left( p\left( s | r \right)\right) = \frac{1}{|s|}  \sum_{i \in s} \left( \sum_{j \in s, j \neq i} \left( \log \left( p\left( v_{j} | v_i, r \right) \right) \right)\right)
\end{equation}
\noindent where $v_i$ is a vertex in the graph and $v_{j}$ are the other vertices in the same group. The probabilities are learned using feature vectors $r_{v_i}$, which are then used as the vertex embeddings. The probability $p\left( v_{j} | v_i, r \right)$ is given by:
\begin{equation} \label{eq:prob_node}
  p\left( v_j | v_i, r \right) = \frac{\exp \left( r\prime _{v_j}^{T} \times r_{v_i} \right) }{ \sum_{v \in V} \left( \exp \left( r\prime _{v}^{T} \times r_{v_i} \right) \right) }
\end{equation}
where $V$ is the set of all nodes and $r\prime ^T_{v_j}$ is transposed output feature vector of vertex $j$ used to make predictions. The representations $r\prime _{v}$ and $r_{v}$ are learned simultaneously by optimizing Equation~\ref{sg_entropy}. This is the same mathematical formulation of SkipGram, and is performed using stochastic gradient ascent with negative sampling \cite{mikolov2013distributed}.

By optimizing this log probability, the algorithm maximizes the log likelihood of predicting a neighbor given a node, creating node embeddings so that nodes with similar neighborhoods have similar representations. Since there is more than one neighbor in each group, this model also makes connected nodes having similar representations, because they will both predict each others neighbors, resulting in representations also with first order similarities. A trade-off between first and second order proximity can be achieved by changing the parameter $k$, which controls the number of nodes within each generated group. Since NBNE considers only the first order neighborhood of a node, it is fast and works well in practice.
\begin{center}
\begin{figure*}
\hspace*{1.2in}
  \includegraphics[width=11cm]{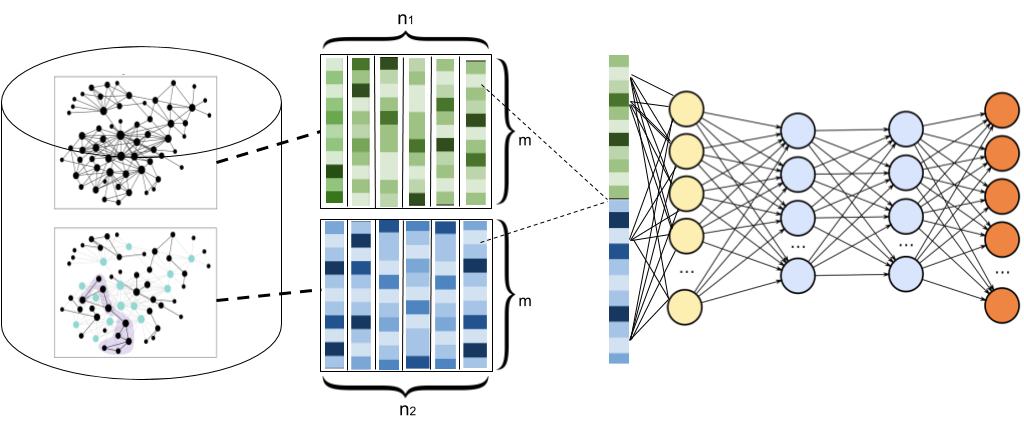}
  \caption{The modeling strategy for multi-graph embedding. The graphs are trained separately and the learned representations form a \begin{math} 2 \times m  \end{math} length vector for each node. The vectors are then fed to feedforward deep neural network in order to find the most probable label.}
\label{fig:modeling_strategy}
\end{figure*}
\end{center}

\subsection{Modeling strategy} \label{subsec:modeling_strategy}
The NBNE is applied on three graphs in order to learn the vector representations of user-user, user-seller and seller-attributes interactions. It is important to mention that it is in fact possible to form a single graph of user, seller and seller attributes and find the representations in a single space, but we train three graphs separately for the following reasons: a) it is easier to interpret the meaning of embeddings for separate graphs than a single graph with complex interactions; b) since graph learning algorithms approximate complete graph exploration through random walks or neighborhood permutations, creating multiple representations for huge graphs can reduce the learning error and generate better embeddings \cite{epasto2019single}; c) different embeddings may be used in different problems, since they have different meanings.

After learning the node representations for user-user, user-seller and seller-attributes, the embeddings are unified in order to create new concepts/vectors. The concatenated vectors are then fed to a  Multilayer Perceptron (MLP) or any other classification algorithm to solve classification or regression problems (see figure \ref{fig:modeling_strategy}). Since vector size is set manually, in order to avoid the dimensionality problem and choosing the most informative set of variables, the best set of features were chosen using feature ranking with recursive feature elimination, Support Vector Machine (SVM) with linear kernel as the estimator and a 5-fold cross validation.

\subsection{Embedding mimic}
The proposed method seeks to learn node embeddings in simpler concepts, such as friendship and buying history, and then utilize different combinations of these representations to model more complex concepts for users with no or limited financial records. Since there is no embedding for new users (until the model is trained with new data) or individuals with no buying history, we propose \textit{Embedding Mimic}, which is the process of generating embeddings for users with no learned representations. Two methods are presented: 
\begin{figure}
\hspace*{0.5in}
  \includegraphics[width=7cm, height = 5cm]{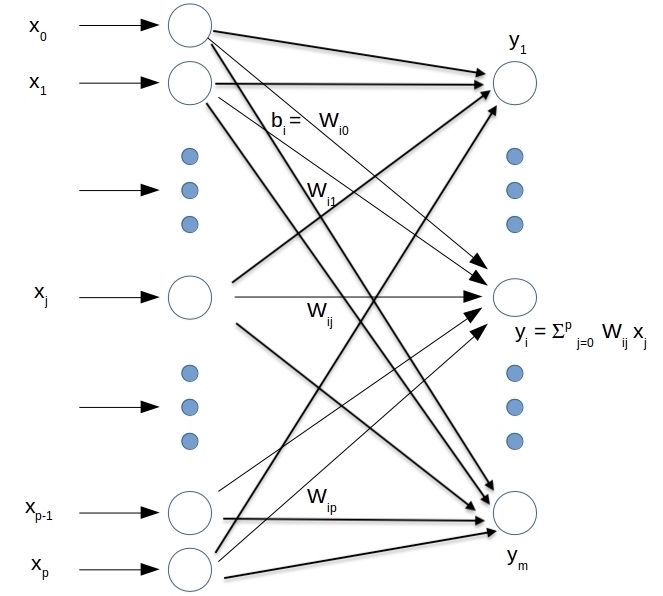}
  \caption{Feed Forward Neural Network can be used to learn the multivariate representations of unseen customers through a regression task.}
\label{fig:multivariate-linear-regression}
\end{figure}

\begin{itemize}
\item The Naïve method is based on recreating new representations by calculating the mean of embedding features of user's direct friends and their buying activities. We can think of the average of node embeddings as a continuous space version of the traditional bag-of-words representation.

\item Multivariate regression as a means of finding a function  \textit{F} which maps the neighboring nodes' embeddings:
\end{itemize}

\begin{equation}
    F: R^{n \times D} \rightarrow R ^ {D}
\end{equation}
\begin{equation}
    F(Neighbor(node, G)) \approx  E(node)
\end{equation}
where \begin{math} n \end{math} and \begin{math} D \end{math} are number of neighbors and embedding dimension respectively.
The function \begin{math} F \end{math} can be approximated by a simple or multiple-hidden-layer neural networks as illustrated in Figure \ref{fig:multivariate-linear-regression} where \begin{math} X_{0}, X_{1}, ..., X_{p} \end{math} are predictors of the concatenated neighborhood embeddings and p is: 

\begin{equation}
    p = n \times D
\end{equation}

The simulated embedding of unseen nodes can then be used in subsequent classification or regression tasks.

\section{Results} \label{sec:res}
The generated embeddings were utilized in two learning tasks related to modeling new user's buying behavior and credit worthiness.

\subsection{Modeling buying behavior}
Buying habits are customers' tendencies to purchase products and hire services. Modeling these tendencies results in having better knowledge about customers and recommending the most relevant products and sellers. Embedding vectors of friendship network of more than 100,000 users and 14000 sellers were learned and then unified as shown in Figure \ref{fig:modeling_strategy}. In order to generate negative samples, complement of graph is generated to find a set the non existent edges (a list of sellers by which the user has not purchased any product). The class label is then set to 1 if user have a history of purchase with a seller and to 0 otherwise. Finally, we randomly split the data into training and test sets following a 80/20 ratio and stratified sampling. No information of users in test set are present in the training set.

Multi-layer Perceptron classifier with learning rate = 0.001, hidden layer size = 100 and max iteration = 200, XGBoost with maximum depth = 3, learning rate = 0.1 and number of estimators = 100, Logistic Regression with penalty \textit{l1} norm and regularization parameter C = 1, and K-NN with \textit{K} = 3 were used to train and test classification performance.

\begin{table}[ht]
  \begin{center}
    \caption{Precision, recall and f1-score for binary classification of Multi-layer Perceptron for buying prediction.}
    \label{tab:table1}
    \begin{tabular}{c|c|c|c} %
      \textbf{Model | class} & \textbf{Precision (\%)} & \textbf{Recall (\%)} & \textbf{f1-score (\%)}\\
      \hline
      MLP | 0 & 80 & 75 & 77\\ \hline
      MLP | 1 & 78 & 81& 79\\ \hline
      XGBoost | 0 & 80 & 78 & 79\\ \hline
      XGBoost | 1 & 75 & 81& 78\\ \hline
      K-NN | 0 & 82 & 74 & 78\\ \hline
      K-NN | 1 & 76 & 83 & 79\\ \hline
    \end{tabular}
  \end{center}
\end{table}

\begin{figure}[ht]
  \includegraphics[width=8cm]{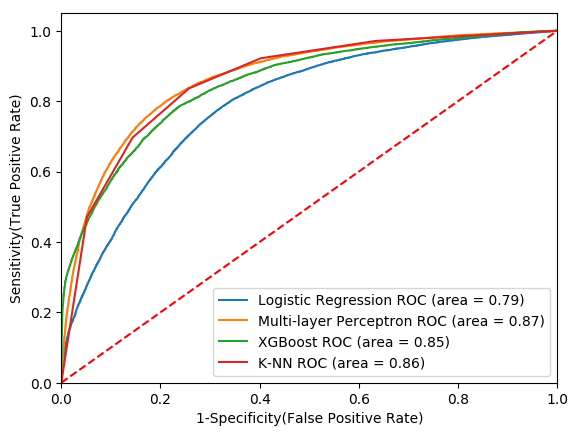}
  \caption{AUC-ROC curve for Logistic Regression, Multi-layer Perceptron, XGBoost and K-NN.}
\label{fig:AUC_purchase}
\end{figure}

As illustrated in Table \ref{tab:table1} and Figure \ref{fig:AUC_purchase}, Multi-layer Perceptron with AUC-ROC score = 87\%, 80\% and 78\% for precision of negative and positive classes, XGBoost with AUC-ROC score = 85\%,  80\% and 75\% of precision for negative and positive classes, and K-nearest Neighbors with  AUC-ROC score = 87\%, 82\% and 76\% precision for positive and negative classes have the best classification results.
\subsection{Modeling credit worthiness}
As mentioned before, machine learning have been considered for modeling the credit risk score. More often, the features that are chosen for analysis not only are not very effective in terms of the collection of relevant data, they are expensive to acquire and to update \cite{kumar2020review}. As a recent study shows, there is a need for new approaches and new way of thinking wherein the machine learning models can be improved to take into account diverse conditions which can analyze the credit profile of the individuals in more diversified manner \cite{kumar2020review}.

Here, we consider using the friendship network and buying habits of friends (and those who have a high probability of making friendship in the future) to predict the credit worthiness of unseen users. Embeddings of more than 205,000 users and their purchase habits were learned and unified and then train and test set are split with 80/20 ratio. Since the data are highly imbalanced (84\% of data is related to users with high risk of credit default and were rejected to receive credit card, and 16\% related to those who received credit card) stratified random sampling was applied in order to preserve the percentage of samples for each class in train and test sets. Because Logistic Regression is one of the most common algorithms in modeling credit worthiness \cite{desai1996comparison} and due to the fact that it can be applied to imbalanced and rare events data, we analysed the results using Logistic Regression with \textit{l1} norm regularization parameter C = 1 as penalty parameter.

\begin{table}[ht]
  \begin{center}
    \caption{Precision, recall and f1-score for binary classification of Logistic Regression for predicting credit worthiness.}
    \label{tab:table2}
    \begin{tabular}{c|c|c|c} %
      \textbf{} & \textbf{Precision (\%)} & \textbf{Recall (\%)} & \textbf{f1-score (\%)}\\
      \hline
       0 & 75 & 88 & 81\\ \hline
       1 & 65 & 43& 52\\ \hline \hline
      macro avg & 70 & 65 & 66\\ \hline
      weighted avg & 71 & 72 & 71\\ \hline
    \end{tabular}
  \end{center}
\end{table}

\begin{figure}[ht]
  \includegraphics[width=8cm]{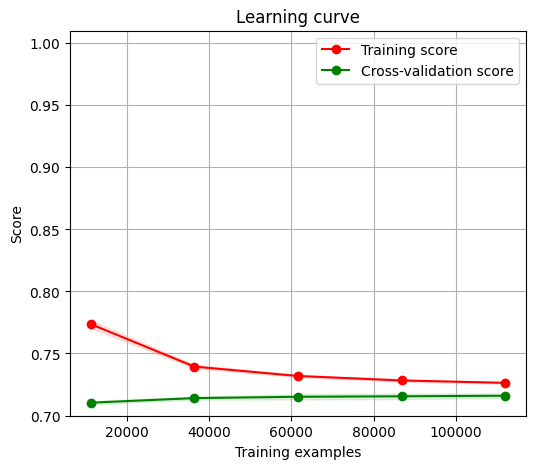}
  \caption{Learning curve of the model for predicting credit worthiness. As seen, cross-validation and training learning curves converge at similar values and the gap between the curves is small.}
  \label{fig:learning_curve}
\end{figure}

As seen in Table \ref{tab:table2}, Logistic Regression with 71\% weighted average precision is able to distinguish those individuals with high and low scores of credit worthiness. If the goal is correctly predicting the positive class (class = 1), by increasing the threshold in favor of losing recall, it is possible to reach up to 75\% (or more) precision. This result is comparable with the results of state of the art methods which use expensive, not always available and hard to update features like socio-economic, financial, occupational and personal variables \cite{kumar2020review}, \cite{singh2017comparative}. As shown in Figure \ref{fig:learning_curve}, both curves converge to a same value and there is an ideal small gap between training and cross validation scores which is an indication of a small variance of the model.

\begin{table}[ht]
  \begin{center}
    \caption{Precision, recall and f1-score for binary classification of Logistic Regression for predicting credit worthiness when only the information of credit worthiness of friends without considering their buying habits is used for prediction.}
    \label{tab:table3}
    \begin{tabular}{c|c|c|c} %
      \textbf{} & \textbf{Precision (\%)} & \textbf{Recall (\%)} & \textbf{f1-score (\%)}\\
      \hline
       0 & 86 & 98 & 92\\ \hline
       1 & 60 & 15& 24\\ \hline \hline
      macro avg & 73 & 57 & 58\\ \hline
      weighted avg & 82 & 85 & 81\\ \hline
    \end{tabular}
  \end{center}
\end{table}

We also tested the possibility of using only information of creditworthiness of friends (or those individuals with high probability of making friendship with a specific user in the future) so as to analyse the effect of buying habits in creditworthiness. As the results of Table \ref{tab:table3} shows, in this case the model is capable of detecting the individuals with low credit worthiness (class 0) with 86\% precision although precision and recall for users with high credit worthiness is decreased by 5\% and 28\%. This indicates that depending on business decisions, both models can be used to identify either individuals with high probability of default or those with high credit worthiness, facilitating credit-granting process.

The above results indicate that it is indeed possible to infer complex characteristics of individuals, like buying behavior and credit worthiness, by utilizing friendship relations in social networks. 

\section{Conclusion} \label{sec:con}
In this study, we examined the feasibility of using multiple graph embeddings instead of expensive, hard to update and sometimes inadequate socio-economic, financial, occupational and personal features to model customer's complex characteristics. A multi-graph embedding approach was proposed to learn different representations of individuals and then to combine in order to create new concepts. The unified embeddings are then fed to machine learning algorithms like Multi-layer Perceptron and Logistic Regression to model a diversity of individuals' social and financial characteristics. Using the current method, it is possible to infer credit worthiness and buying behavior of individuals only by observing financial and purchase behavior of nearby individuals in the embedding space. 

\section{Conflict of Interest}
The current method was proposed and tested by a group of data scientists from PicPay. Any opinions, findings, and conclusions expressed in this manuscript are those of the authors and do not necessarily reflect the views, official policy or position of PicPay.

\bibliography{refs}{}

\bibliographystyle{IEEEtran}

\vspace{12pt}

\end{document}